\documentclass[letterpaper, 10 pt, conference]{ieeeconf}  

\IEEEoverridecommandlockouts                              

\usepackage{graphicx,amsbsy,amssymb,url}
\usepackage{amsmath}
\usepackage{thmtools}
\usepackage{booktabs}
\usepackage{comment,dsfont}
\usepackage{epstopdf}
\usepackage{cite}
\usepackage{amsmath}
\usepackage[textsize=scriptsize]{todonotes}
\usepackage{mathtools}
\usepackage[noend]{algpseudocode}
\usepackage{algorithm}
\usepackage{float}
\usepackage{tikz}
\usetikzlibrary{positioning,plotmarks,calc}
\usepackage{pgfplots}
\pgfplotsset{compat=1.9}

\newtheorem{remark}{Remark}
\newtheorem{definition}{Definition}
\newtheorem{theorem}{Theorem}
\newtheorem{corollary}{Corollary}

\newcommand{\R}{\ensuremath{\mathbb{R}}}

\DeclareMathOperator*{\argmin}{arg\,min}

\DeclarePairedDelimiter{\parens}{\lparen}{\rparen}
\DeclarePairedDelimiter{\abs}{\lvert}{\rvert}
\DeclarePairedDelimiter{\norm}{\lVert}{\rVert}
\DeclarePairedDelimiter{\ceil}{\lceil}{\rceil}
\DeclarePairedDelimiter{\floor}{\lfloor}{\rfloor}

\makeatletter
\let\oldparens\parens
\def\parens{\@ifstar{\oldparens}{\oldparens*}}
\let\oldnorm\norm
\def\norm{\@ifstar{\oldnorm}{\oldnorm*}}
\let\oldceil\ceil
\def\ceil{\@ifstar{\oldceil}{\oldceil*}}
\let\oldfloor\floor
\def\floor{\@ifstar{\oldfloor}{\oldfloor*}}
\let\oldabs\abs
\def\abs{\@ifstar{\oldabs}{\oldabs*}}
\makeatother

\title{\LARGE \bf
Algorithmic Aspects of Inverse Problems Using Generative Models
}

\author{Chinmay Hegde$^{1}$
\thanks{$^{1}$CH is with the Electrical and Computer Engineering Department at
   Iowa State University, Ames, IA USA.
     Email: chinmay@iastate.edu.}%
}

\begin{document}

\maketitle
\thispagestyle{empty}
\pagestyle{empty}

\begin{abstract}

The traditional approach of hand-crafting priors (such as
      sparsity) for solving inverse problems is slowly being
      replaced by the use of richer learned priors (such as those
      modeled by generative adversarial networks, or GANs). In
      this work, we study the algorithmic aspects of such a
      learning-based approach from a theoretical perspective. For
      certain generative network architectures, we establish a
      simple non-convex algorithmic approach that (a) theoretically
      enjoys linear convergence guarantees for certain inverse problems, 
      and (b) empirically improves upon conventional
      techniques such as back-propagation. We also propose an extension of our approach that can handle model mismatch (i.e., situations where the generative network prior is not exactly applicable.) Together, our contributions serve as building blocks towards a more complete algorithmic understanding of generative models in inverse problems.

\end{abstract}

\section{INTRODUCTION}

\subsection{Motivation}

Inverse problems arise in a diverse range of application domains including computational imaging, optics, astrophysics, and seismic geo-exploration. In each of these applications, there is a target signal or image (or some other quantity of interest) to be obtained; a device (or some other physical process) records measurements of the target; and the goal is to reconstruct an estimate of the signal from the observations. 

%
Let us suppose that $x \in \R^n$ denotes the signal of interest and $y = \mathcal{A}(x) \in \R^m$ denotes the observed measurements. When $m < n$ the inverse problem is ill-posed, and some kind of prior (or regularizer) is necessary to obtain a meaningful solution. A common technique used to solve ill-posed inverse problems is to solve a constrained optimization problem:
\begin{align}
\widehat{x} &= \argmin~F(x),~~\label{eq:cop}\\
&\text{s. t.}~~~x \in \mathcal{S},\nonumber
\end{align}
where $F$ is an objective function that typically depends on $y$ and $\mathcal{A}$, and $\mathcal{S} \subseteq \R^n$ captures some sort of \emph{structure} that $x$ is assumed to obey. 

A very common modeling assumption, particularly in signal and image processing applications, is \emph{sparsity}, wherein $\mathcal{S}$ is the set of sparse vectors in some (known) basis representation. The now-popular framework of \emph{compressive sensing} studies the special case where the forward measurement operator $\mathcal{A}$ can be modeled as a linear operator that satisfies certain (restricted) stability properties; when this is the case, accurate estimation of $x^*$ can be performed provided the signal $x$ is sufficiently sparse~\cite{candes2006compressive}. Parallel to the development of algorithms that leverage sparsity priors, the last decade has witnessed analogous approaches for other families of structural constraints. These include structured sparsity~\cite{modelcs,surveyEATCS}, union-of-subspaces~\cite{MarcoCISS}, dictionary models~\cite{elad2006image,aharon2006rm}, total variation models~\cite{chan2006total}, analytical transforms~\cite{sairprasad}, among many others. 

Lately, there has been renewed interest in prior models that are parametrically defined in terms of a \emph{deep neural network}. We call these \emph{generative network} models. Specifically, we define 
\[
\mathcal{S} = \{x \in \R^n~|~x  = G(z),~z \in \R^k \}
\] 
where $z$ is a $k$-dimensional latent parameter vector and $G$ is parameterized by the weights and biases of a $d$-layer neural network. One way to obtain such a model is to train a generative adversarial network (GAN)~\cite{goodfellow2014generative}. A well-trained GAN closely captures the notion of a signal (or image) being `natural'~\cite{berthelot2017began}, leading many to speculate that the range of such generative models can approximate a low-manifold containing naturally occurring images. Indeed, GAN-based neural network learning algorithms have been successfully employed to solve \emph{linear} inverse problems such as image super-resolution and inpainting~\cite{yeh2016semantic, ledig2016photo}. However, most of these approaches are heuristic, and a general theoretical framework for analyzing the performance of such approaches is not available at the moment.

\subsection{Contributions}

Our focus in this paper is to take some initial steps into building such a theoretical framework. Specifically, we wish to understand the \emph{algorithmic} costs involving in using generative network models for inverse problems: how computationally challenging they are, whether they provably succeed, and how to make such models robust.

The starting point of our work is the recent, seminal paper by \cite{bora2017compressed}, who study the benefits of using generative models in the context of compressive sensing. In this paper, the authors pose the estimated target as the solution to a non-convex optimization problem and establish upper bounds on the \emph{statistical} complexity of obtaining a ``good enough'' solution. Specifically, they prove that if the generative network is a mapping $G : \R^k \rightarrow \R^n$ simulated by a $d$-layer neural network with width $\leq n$ and with activation functions obeying certain properties, then $m = O(kd \log n)$ random observations are sufficient to obtain a good enough reconstruction estimate. However, the authors do not study the \emph{algorithmic} costs of solving the optimization problem, and standard results in non-convex optimization are sufficient to only obtain sublinear convergence rates. In earlier work~\cite{ganICASSP}, we established an algorithm with linear convergence rates for the same (compressive sensing) setup, and demonstrated its empirical benefits.

However, the earlier work~\cite{ganICASSP} only provided an algorothm (and analysis) for linear inverse problems. In this work, we generalize this to a much wider range of \emph{nonlinear} inverse problems. Using standard techniques, we propose a generic algorithm for solving~\eqref{eq:cop}, analyze its performance, and prove that it demonstrates linear convergence. This constitutes \textbf{Contribution I} of this paper. 

A drawback of \cite{bora2017compressed} (and our subsequent work~\cite{ganICASSP}) is the inability to deal with targets that are outside the range of the generative network model. This is not merely an artifact of their analysis; generative networks are rigid in the sense that once they are learned, they are incapable of reproducing any target outside their range. (This is in contrast with other popular parametric models such as sparsity models; these exhibit a ``graceful decay'' property in the sense that if the sparsity parameter $s$ is large enough, such models capture all possible points in the target space.) This issue is addressed, and empirically resolved, in the recent work of~\cite{sparsegen} who propose a hybrid model combining both generative networks and sparsity. This leads to a non-convex optimization framework (called \emph{SparseGen}) which the authors theoretically analyze to obtain analogous statistical complexity results. However, here too, the theoretical contribution is primarily statistical and the algorithmic aspects of their setup are not discussed. 

We address this gap, and propose an alternative algorithm for this framework. Our algorithm is new, and is a nonlinear extension of our previous work~\cite{spinisit,spinIT}. Under (fairly) standard assumptions, this algorithm also can be shown to demonstrate linear convergence. This constitutes \textbf{Contribution II} of this paper. 

In summary: we complement the work of \cite{bora2017compressed} and \cite{sparsegen} by providing algorithmic upper bounds for the corresponding problems that are studied in those works. Together, our contributions serve as further building blocks towards an algorithmic theory of generative models in inverse problems.

\subsection{Techniques}

At a high level, our algorithms are standard. The primary novelty is in their applications to generative network models, and some aspects of their theoretical analysis.

Suppose that $G : \R^k \rightarrow \R^n$ is the generative network model under consideration. The cornerstone of our analysis is the assumption of an $\varepsilon$-\emph{approximate} (Euclidean) projection oracle onto the range of $G$. In words, we pre-suppose the availability of a computational routine $P_G$ that, given any vector $x \in \R^n$, can return a vector $x' \in \text{Range}(G)$ that approximately minimizes $\norm{x - x}_2^2$.  The availability of this oracle, of course, depends on $G$ and we comment on how to construct such oracles below in Section~\ref{sec:conc}.

The first algorithm (for solving~\eqref{eq:cop}) is a straightforward application of \emph{projected gradient descent}, and is a direct nonlinear generalization of the approach proposed in~\cite{ganICASSP}. The main difficulty is in analyzing the algorithm and proving linear convergence. To show this, we assume that the objective function in~\eqref{eq:cop} obeys the Restricted Strong Convexity/Smoothness assumptions~\cite{raskutti2010restricted}. With this assumption, proof of convergence follows from a straightforward modification of the proof given in~\cite{jainkar2017}.

The second algorithm (for handling model mismatch in the target) is new. The main idea (following the lead of \cite{sparsegen}) is to pose the target $x$ as the superposition of two components: $x = G(z) + \nu$, where $\nu$ can be viewed as an ``innovation'' term that is $s$-sparse in some fixed, known basis $B$. The goal is now to recover both $G(z)$ and $\nu$. This is reminiscent of the problem of source separation or signal demixing~\cite{mccoyTropp2014}, and in our previous work~\cite{spinIT,NLDemix_TSP} we proposed greedy iterative algorithms for solving such demixing problems. We extend this work by proving a nonlinear extension, together with a new analysis, of the algorithm proposed in~\cite{spinIT}.  

\section{BACKGROUND AND RELATED WORK}

\begin{figure}
\begin{center}

\tikzstyle{startstop} = [rectangle, rounded corners, minimum height=1cm, text centered, draw=black, fill=red!30]
\tikzstyle{io} = [rectangle, rounded corners, minimum height=1cm, text centered, draw=black, fill=blue!30]
\tikzstyle{process} = [rectangle, rounded corners, minimum height=1cm, text centered, draw=black, fill=orange!30]
\tikzstyle{decision} = [diamond, minimum height=1cm, text centered, draw=black, fill=green!30]
\tikzstyle{arrow} = [thick,->,>=stealth]

\begin{tikzpicture}[node distance=3cm]

\node (start) [startstop] {Image/signal};
\node (in1) [io, right of=start] {Forward model};
\node (pro1) [process, right of=in1] {Observations};

\draw [arrow] (start) -- (in1);
\draw [arrow] (in1) -- (pro1);
\end{tikzpicture}
\end{center}
\caption{\label{fig:ill} Block diagram for a generic inverse problem. The goal is to reconstruct (an estimate of) the data/signal given knowledge of the observations or measurements.}
\end{figure}
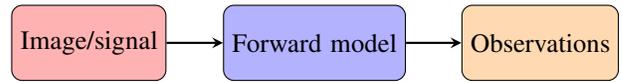


\subsection{Inverse problems}

The study of solving inverse problems has a long history. As discussed above, the general approach to solve an ill-posed inverse problem of the form depicted in Figure~\ref{fig:ill} is to assumes that the target signal/image obeys a \emph{prior}. Older works mainly used hand-crafted signal priors; for example,  \cite{donoho1995noising, xu2010image, dong2011image} employ sparsity priors, and applied them in \emph{linear} inverse problems such as super-resolution, denoising, compressive sensing, and missing entry interpolation. 

\subsection{Neural network models}

The last few years have witnessed the emergence of trained \emph{neural networks} for solving such problems. The main idea is to eschew hand-crafting any priors, and instead \emph{learn} an end-to-end mapping from the measurement space to the image space. This mapping is simulated via a deep neural network, whose weights are learned from a large dataset of input-output training examples  \cite{lecun2015deep}. The works \cite{kulkarni2016reconnet,mousavi2015deep,mousavi2017learning,xu2014deep, dong2016image,kim2016accurate,yeh2017semantic} have used this approach to solve several types of (linear) inverse problems, and has met with considerable success. However, the major limitations are that a new network has to be trained for each new linear inverse problem; moreover, most of these methods lack concrete theoretical guarantees. An exception of this line of work is the powerful framework of~\cite{rick2017one}, which does \emph{not} require retraining for each new problem; however, this too is not accompanied by theoretical analysis of statistical and computational costs.

\subsection{Generative networks}

A special class of neural networks that attempt to directly model the distribution of the input training samples are known as generative  adversarial training networks, or GANs \cite{goodfellow2014generative}. GANs have been shown to provide visually striking results \cite{arjovsky2017wasserstein,pix2pix,berthelot2017began,simonyan2018}. The use of GANs to solve linear inverse problems was advocated in \cite{bora2017compressed}. Specifically, given (noisy) linear observations $y = Ax + e$ of a signal $x \in \R^n$, assuming that $x$ belongs to the range of a generative network $G : \R^n \rightarrow \R^k$, this approach constructs the reconstructed estimate $\hat{x}$ as follows:
\[
\hat{z} = \arg \min_{z \in \R^k} \norm{y - A G(z)}_2^2,~~\hat{x} = G(\hat{z})
\]
If the observation matrix $A \in \R^{m \times n}$ comprises 
$
m = O(kd \log n)
$
i.i.d. Gaussian measurements, then together with regularity assumptions on the generative network, they prove that the solution $\hat{x}$ satisfies:
$$\norm{x - \hat{x}}_2 \leq C \norm{e}_2.$$
for some constant $C$ that can be reliably upper-bounded. In particular, in the absence of noise the recovery of $x$ is exact. However, there is no discussion of how computationally expensive this procedure is. Observe that the above minimization is highly non-convex (since for any reasonable neural network, $G$ is a non-convex function) and possibly also non-smooth (if the neural network contains non-smooth activation functions, such as rectified linear units, or ReLUs). More recently,~\cite{deepprior} improve upon the approach in~\cite{bora2017compressed} for solving more general nonlinear inverse problems (in particular, any inverse problem which has a computable derivative). Their approach involves simultaneously solving the inverse problem and training the network parameters; however, the treatment here is mostly empirical and a theoretical analysis is not provided.

Under similar statistical assumptions as~\cite{bora2017compressed}, the work of~\cite{ganICASSP} provably establishes a linear convergence rate, provided that a projection oracle (on to the range of $G$) is available, but only for the special case of compressive sensing. Our first main result (Contribution I) extends this algorithm (and analysis) to more general nonlinear inverse problems.

\subsection{Model mismatch}

A limitation of most generative network models is that they can only reproduce estimates that are within their range; adding more observations or tweaking algorithmic parameters are completely ineffective if a generative network model is presented with a target that is far away from the range of the model. To resolve this type of model mismatch, the authors of \cite{sparsegen} propose to model the signal $x$ as the superposition of two components: a ``base'' signal $u = G(z)$, and an ``innovation'' signal $v = B \nu$, where $B$ is a known ortho-basis and $\nu$ is an $l$-sparse vector.  In the context of compressive sensing, the authors of \cite{sparsegen} solve a sparsity-regularized loss minimization problem:
\[
(\hat{z}, \hat{v}) = \arg \min_{z, v} \norm{B^T v}_1 + \lambda \norm{y - A(G(z) + v)}_2^2 .
\]
and prove that the reconstructed estimate $\hat{x} = G(\hat{z}) + \hat{v}$ is close enough to $x$ provided $m = O((k + l)d \log n)$ measurements are sufficient. However, as before, the algorithmic costs of solving the above problem are not discussed. Our second main result (Contribution II) proposes a new algorithm for dealing with model mismatches in generative network modeling, together with an analysis of its convergence and iteration complexity.

\section{MAIN ALGORITHM AND ANALYSIS}

Let us first establish some notational conventions. Below, $\norm{\cdot}$ will denote the Euclidean norm unless explicitly specified. We use $O(\cdot)$-notation in several places in order to avoid duplication of constants.  

We use $F(\cdot)$ to denote a (scalar) objective function. We assume that $F$ has a continuous gradient $\nabla F = \left(\frac{\partial F}{\partial x_i}\right)_{i=1}^n$ which can be evaluated at any point $x \in \R^n$.

\subsection{Definitions and assumptions}

We now present certain definitions that will be useful for our algorithms and analysis.

\begin{definition}[Approximate projection] A function $P_G : \R^n \rightarrow \text{Range}(G)$ is an $\varepsilon$-approximate projection oracle if for all $x \in \R^n$, $P_G(x)$ obeys:
\[
\norm{x - P_G(x)}^2_2 \leq \min_{z \in \mathbb{R}^k} \norm{x - G(z)}^2_2 + \varepsilon .
\]
We will assume that for any given generative network $G$ of interest, such a function $P_G$ exists and is computationally tractable\footnote{This may be a very strong assumption, but at the moment we do not know how to relax this. Indeed, the computational complexity of our proposed algorithms are directly proportional to the complexity of such a projection oracle.}. Here, $\varepsilon > 0$ is a parameter that is known \emph{a priori}. 
\end{definition}

\begin{definition}[Restricted Strong Convexity/Smoothness]
Assume that $F$ satisfies $\forall x, y \in S$:
\[
\frac{\alpha}{2} \norm{x - y}_2^2 \leq F(y) - F(x) - \langle \nabla F(x), y - x \rangle \leq \frac{\beta}{2} \norm{x - y}_2^2 .
\]
for positive constants $\alpha, \beta$.
\end{definition}
\medskip
This assumption is by now standard; see~\cite{raskutti2010restricted,jainkar2017} for in-depth discussions. This means that the objective function is strongly convex / strongly smooth along certain directions in the parameter space (in particular, those restricted to the set $S$ of interest). The parameter $\alpha > 0$ is called the restricted strong convexity (RSC) constant, while the parameter $\beta > 0$ is called the restricted strong smoothness (RSS) constant. Clearly, $\beta \geq \alpha$. In fact, throughout the paper, we assume that 
$
1 \leq \frac{\beta}{\alpha} < 2,
$
which is a fairly stringent assumption but again, one that we do not know at the moment how to relax.

\begin{definition}[Incoherence]
A basis $B$ and $\text{Range}(G)$ are called $\mu$-incoherent if for all $u, u' \in \text{Range}(G)$ and all $v, v' \in \text{Span}(B)$, we have:
\[
| \langle u - u', v - v' \rangle | \leq \mu \norm{u - u'}_2 \norm{v - v'}_2.
\]
for some parameter $0 < \mu< 1$.
\end{definition}

\begin{remark}
In addition to the above, we will make the following assumptions in order to aid the analysis. Below, $\gamma$ and $\Delta$ are positive constants.
\begin{itemize}
\item $\norm{\nabla F(x^*)}_2 \leq \gamma$ (gradient at the minimizer is small).
\item $\text{diam}(\text{Range}(G)) = \Delta$ (range of $G$ is compact). 
\item $\gamma \Delta \leq O(\varepsilon)$. 
\end{itemize}
\end{remark}

\begin{algorithm}[t]
	\caption{$\varepsilon$-\textsc{PGD}}
	\label{alg:PGD}
	\begin{algorithmic}[1]
	\State \textbf{Inputs:} $y$, $T$, $\nabla$; \textbf{Output:}  $\widehat{x}$
	\State $x_0 \leftarrow \textbf{0}$ \hspace{17.8em} 
	\While {$t < T$}
	\State $z_t \leftarrow x_t - \eta \nabla F(x_t)$ \hspace{8.8em} 
	\State $x_{t+1} \leftarrow P_G(z_t) $ \hspace{0.6em} 
	\State $t \leftarrow t+1$
	\EndWhile
	\State $\widehat{x} \leftarrow x_{T}$
	\end{algorithmic}
\end{algorithm}

 \subsection{Contribution I: An algorithm for nonlinear inverse problems using generative networks}

We now present our first main result. Recall that we wish to solve the problem:
\begin{align}
\widehat{x} &= \argmin~F(x),~~\label{eq:cop0}\\
&\text{s. t.}~~~x \in \text{Range}(G),\nonumber
\end{align}
where $G$ is a generative network. To do so, we employ \emph{projected gradient descent} using the $\varepsilon$-approximate projection oracle for $G$. The algorithm is described in Alg. \ref{alg:PGD}. We obtain the following theoretical result:

\begin{theorem}
\label{thm:pgd}
If $F$ satisfies RSC/RSS over $\text{Range}(G)$ with constants $\alpha$ and $\beta$, then $\varepsilon$-PGD (Alg.\ \ref{alg:PGD}) convergences linearly up to a ball of radius $O(\gamma \Delta) \approx O(\varepsilon)$. 
\[
F(x_{t+1}) - F(x^*) \leq \left(2 - \frac{\beta}{\alpha}\right) (F(x_t) - F(x^*)) + O(\varepsilon) \, .
\]
\end{theorem}
\proof
The proof is a minor modification of that in \cite{jainkar2017}. For simplicity we will assume that $\norm{\cdot}$ refers to the Euclidean norm. Let us suppose that the step size $\eta = \frac{1}{\beta}$. Define
\[
z_t = x_t - \eta \nabla F(x_t) .
\]
By invoking RSS, we get:
\begin{align*}
& F(x_{t+1}) - F(x_t) \\
& \leq \langle \nabla F(x_t) , x_{t+1} - x_t \rangle + \frac{\beta}{2} \norm{x_{t+1} - x_t}^2 \\
&=  \frac{1}{\eta} \langle x_t - z_t, x_{t+1} - x_t \rangle + \frac{\beta}{2} \norm{x_{t+1} - x_t}^2 \\
&= \frac{\beta}{2} \left( \norm{x_{t+1} - x_t}^2 + 2 \langle x_t - z_t,  x_{t+1} - x_t \rangle + \norm{x_t - z_t}^2 \right) \\
&~~~- \frac{\beta}{2} \norm{x_t - z_t}^2 \\
&= \frac{\beta}{2} \left(\norm{x_{t+1} - z_t}^2 - \norm{x_t - z_t}^2 \right),
\end{align*}
where the last few steps are consequences of straightforward algebraic manipulation. 

Now, since $x_{t+1}$ is an $\varepsilon$-approximate projection of $z_t$ onto $\text{Range}(G)$ and $x^* \in \text{Range}(G)$, we have:
\[
\norm{x_{t+1} - z_t}^2 \leq \norm{x^* - z_t}^2 + \varepsilon.
\]
Therefore, we get:
\begin{align*}
& F(x_{t+1}) - F(x_t) \\
&\leq \frac{\beta}{2} \left(\norm{x^* - z_t}^2 - \norm{x_t - z_t}^2 \right) + \frac{\beta \varepsilon}{2} \\
&=\frac{\beta}{2} \left(\norm{x^* - x_t + \eta \nabla F(x_t)}^2 - \norm{\eta \nabla F(x_t)}^2 \right) + \frac{\beta \varepsilon}{2} \\
&= \frac{\beta}{2} \left( \norm{x^* - x_t}^2 + 2 \eta \langle x^* - x_t, \nabla F(x_t) \rangle \right) + \frac{\beta \varepsilon}{2} \\
&= \frac{\beta}{2} \norm{x^* - x_t}^2 + \langle x^* - x_t, \nabla F(x_t) \rangle + \frac{\beta \varepsilon}{2}.
\end{align*}
However, due to RSC, we have:
\begin{align*}
\frac{\alpha}{2} \norm{x^* - x_t}^2 &\leq F(x^*) - F(x_t) - \langle x^* - x_t, \nabla F(x_t) \rangle, \\
\langle x^* - x_t, \nabla F(x_t) \rangle &\leq F(x^*) - F(x_t) - \frac{\alpha}{2} \norm{x^* - x_t}^2 .
\end{align*}
Therefore,
\begin{align*}
& F(x_{t+1}) - F(x_t) \\
&\leq \frac{\beta - \alpha}{2} \norm{x^* - x_t}^2 + F(x^*) - F(x_t) + \frac{\beta \varepsilon}{2} \\
&\leq \frac{\beta - \alpha}{2} \cdot \frac{2}{\alpha} \left( F(x_t) - F(x^*) - \langle x_t - x^*, \nabla F(x^*) \rangle \right)  \\
& ~~~~+ F(x^*) - F(x_t) + \frac{\beta \varepsilon}{2} \\
&\leq \left(2 - \frac{\beta}{\alpha}\right) \left( F(x^*) - F(x_t) \right) + \frac{\beta - \alpha}{\alpha} \gamma \Delta + \frac{\beta \varepsilon}{2} ,
\end{align*}
where the last inequality follows from Cauchy-Schwartz and the assumptions on $\norm{\nabla F(x^*)}$ and the diameter of $\text{Range}(G)$.  Further, by assumption, $\gamma \Delta \leq O(\varepsilon)$. Rearranging terms, we get:
\[
F(x_{t+1}) - F(x^*) \leq \left(\frac{\beta}{\alpha} - 1\right) \left( F(x_t) - F(x^*) \right) + C \varepsilon .
\]
for some constant $C > 0$.
\endproof

This theorem asserts that the distance between the objective function at any iteration to the optimum \emph{decreases by a constant factor} in every iteration. (The decay factor is $\frac{\beta}{\alpha} - 1$, which by assumption is a number between 0 and 1). Therefore, we immediately obtain linear convergence of $\varepsilon$-PGD up to a ball of radius $O(\varepsilon)$:

\begin{corollary}
After $T = O(\log \frac{F(x_0) - F(x^*)}{\varepsilon})$ iterations, $F(x_T) \leq F(x^*) + O(\varepsilon$) .
\end{corollary}

Therefore, the overall running time can be bounded as follows:
\[
\text{Runtime} \leq (T_{\varepsilon-\textsc{Proj}} + T_\nabla) \times \log (1 / \varepsilon ) .
\]
See~\cite{ganICASSP} for empirical evaluations of PGD applied to a linear inverse problem (compressed sensing recovery).

\subsection{Contribution II: Addressing signal model mismatch}

We now generalize the $\varepsilon$-PGD algorithm to handle situations involving signal model mismatch. Assume that the target signal can be decomposed as:
\[
x = G(z) + v,
\]
where $\norm{B^T v}_0 \leq l \ll n$ for some ortho-basis $B$. 

For this model, we attempt to solve a (slightly) different optimization problem:
\begin{align}
\widehat{x} &= \argmin~F(x),~~\label{eq:mcop}\\
&\text{s. t.}~~~x = G(z) + v,,\nonumber \\
&~~~~~~~~\norm{B^T v}_0 \leq l.
\end{align}

We propose a new algorithm to solve this problem that we call \emph{Myoptic $\varepsilon$-PGD}. This algorithm is given in Alg.\ \ref{alg:mPGD}\footnote{The algorithm is a variant of block-coordinate descent, except that the block updates share the same gradient term.}.

\begin{algorithm}[t]
	\caption{\textsc{Myopic $\varepsilon$-PGD}}
	\label{alg:mPGD}
	\begin{algorithmic}[1]
	\State \textbf{Inputs:} $y$, $T$, $\nabla$; \textbf{Output:}  $\widehat{x}$
	\State $x_0, u_0, v_0 \leftarrow \textbf{0}$ \hspace{17.8em} 
	\While {$t < T$}
	\State $u_{t+1} = P_G(u_t - \eta \nabla_x F(x_t))$
	\State $v_{t+1} = \text{Thresh}_{B,l}(v_t - \eta \nabla_x F(x_t))$
	\State $x_{t+1} = u_{t+1} + \nu_{t+1}$
	\State $t \leftarrow t+1$
	\EndWhile
	\State $\widehat{x} \leftarrow x_{T}$
	\end{algorithmic}
\end{algorithm}

\begin{theorem}
\label{thm:mpgd}
Let $\oplus$ denote the Minkowski sum. If $F$ satisfies RSC/RSS over $\text{Range}(G) \oplus \text{Span}(B)$ with constants $\alpha$ and $\beta$, and if we assume $\mu$-incoherence between $B$ and $\text{Range}(G)$, we have:
\[
F(x_{t+1}) - F(x^*) \leq \left(\frac{2 - \frac{\beta}{\alpha} \frac{1 - 2.5 \mu}{1 - \mu}}{1 - \frac{\beta}{2\alpha} \frac{\mu}{1 - \mu}}\right) (F(x_t) - F(x^*)) + O(\varepsilon) \, .
\]
\end{theorem}
\bigskip
\proof
We will generalize the proof technique of~\cite{spinIT}.
We first define some auxiliary variables that help us with the proof. Let:
\begin{align*}
z_t &= x_t - \eta \nabla F(x_t), \\
z_t^u &= u_t - \eta \nabla F(x_t), \\
z_t^v &= v_t - \eta \nabla F(x_t) .
\end{align*}
and let $x^* = u^* + v^*$ be the minimizer that we seek. 
As above, by invoking RSS and with some algebra, we obtain:
\begin{align}
F(x_{t+1}) - F(x^*) \leq \frac{\beta}{2} \left(\norm{x_{t+1} - z_t}^2 - \norm{x_t - z_t}^2 \right),
\label{eq:upperb_rsc}
\end{align}
However, by definition, 
\begin{align*}
x_{t+1} &= u_{t+1} + v_{t+1}, \\
x_{t} &= u_{t} + v_{t} .
\end{align*}
Therefore, 
\begin{align*}
&\norm{x_{t+1} - z_t}^2  \\
& = \| u_{t+1} - (u_t - \eta \nabla F(x_t)) + \\
&~~~~v_{t+1} - (v_t - \eta \nabla F(x_t)) + \eta \nabla F(x_t) \|^2 \\
& = \| u_{t+1} - (u_t - \eta \nabla F(x_t)) \|^2 + \| \eta \nabla F(x_t) \|^2 +\\ 
&~~~~ \| v_{t+1} - (v_t - \eta \nabla F(x_t)) \|^2 \\
&~~+ 2 \langle u_{t+1} - (u_t - \eta \nabla F(x_t)), \eta \nabla F(x_t) \rangle \\
&~~+ 2 \langle v_{t+1} - (v_t - \eta \nabla F(x_t)), \eta \nabla F(x_t) \rangle \\
&~~+ 2 \langle u_{t+1} - (u_t - \eta \nabla F(x_t)), v_{t+1} - (v_t - \eta \nabla F(x_t)) \rangle .
\end{align*}
But $u_{t+1}$ is an $\varepsilon$-projection of $z_t^u$ and $u^*$ is in the range of $G$, we have: 
\[
\norm{u_{t+1} - z_t^u}^2 \leq \norm{u^* - z^u_t}^2 + \varepsilon.
\]
Similarly, since $v_{t+1}$ is an $l$-sparse thresholded version of $z_t^v$, we have:
\[
\norm{v_{t+1} - z_t^v}^2 \leq \norm{v^* - z^v_t}^2 .
\]
Plugging in these two upper bounds, we get:
\begin{align*}
&\norm{x_{t+1} - z_t}^2  \\
& \leq \| u^* - (u_t - \eta \nabla F(x_t)) \|^2 + \varepsilon \\ 
&~~~~ \| \eta \nabla F(x_t) \|^2 + \| v^* - (v_t - \eta \nabla F(x_t)) \|^2 \\
&~~+ 2 \langle u_{t+1} - (u_t - \eta \nabla F(x_t)), \eta \nabla F(x_t) \rangle \\
&~~+ 2 \langle v_{t+1} - (v_t - \eta \nabla F(x_t)), \eta \nabla F(x_t) \rangle \\
&~~+ 2 \langle u_{t+1} - (u_t - \eta \nabla F(x_t)), v_{t+1} - (v_t - \eta \nabla F(x_t)) \rangle .
\end{align*}
Expanding squares and cancelling (several) terms, the right hand side of the above inequality can be simplified to obtain:
\begin{align*}
&\norm{x_{t+1} - z_t}^2 \\ 
&\leq \norm{u^* + v^* - z_t}^2 +  \varepsilon  \\
&~~~~ + 2 \langle u_{t+1} - u_t, v_{t+1} - v_t \rangle - 2 \langle u^* - u_t, v^* - v_t \rangle \\
&= \norm{x^* - z_t}^2  + \varepsilon + 2 \langle u_{t+1} - u_t, v_{t+1} - v_t \rangle \\
&~~~~- 2 \langle u^* - u_t, v^* - v_t \rangle .
\end{align*}
Plugging this into \eqref{eq:upperb_rsc}, we get:
\begin{align*}
& F(x_{t+1}) - F(x^*) \\
& \leq \underbrace{\frac{\beta}{2} \left( \norm{x^* - z_t}^2 - \norm{x_t - z_t}^2 \right)}_{\mathbb{T}_1}  \\
&~~+ \underbrace{\beta \left( \langle u_{t+1} - u_t, v_{t+1} - v_t \rangle - 2 \langle u^* - u_t, v^* - v_t \rangle  \right)}_{\mathbb{T}_2} \\
&~~+ \frac{\beta \varepsilon}{2}.
 \end{align*}
We already know how to bound the first term $\mathbb{T}_1$, using an identical argument as in the proof of Theorem~\ref{thm:pgd}. We get:
\[
\mathbb{T}_1 \leq \left(2 - \frac{\beta}{\alpha}\right) \left( F(x^*) - F(x_t) \right) + \frac{\beta - \alpha}{\alpha} \gamma \Delta .
\]
The second term $\mathbb{T}_2$ can be bounded as follows. First, observe that
\begin{align*}
& | \langle u_{t+1} - u_t, v_{t+1} \rangle | \\
&\leq \mu \norm{u_{t+1} - u_t} \norm{v_{t+1} - v_t} \\
&\leq \frac{\mu}{2} \left( \norm{u_{t+1} - u_t}^2 + \norm{v_{t+1} - v_t}^2 \right) \\
&\leq \frac{\mu}{2} \left( \norm{u_{t+1} + v_{t+1} - u_t - v_t}^2 \right) \\
&~~~~+ \mu | \langle u_{t+1} - u_t, v_{t+1} - v_t \rangle | .
\end{align*}
This gives us the following inequalities:
\begin{align*}
& | \langle u_{t+1} - u_t, v_{t+1} \rangle | \\
&\leq \frac{\mu}{2(1 - \mu)} \norm{x_{t+1} - x_t}^2 \\
&= \frac{\mu}{2(1 - \mu)} \left(  \norm{x_{t+1} - x^*}^2 +  \norm{x_{t} - x^*}^2 + \right. \\
& \left. ~~~~~~~~~~~~~~~~~~~~2 | \langle x_{t+1} - x^*, x_t - x^* \rangle  |  \right) \\
&\leq \frac{\mu}{1 - \mu} \left( \norm{x_{t+1} - x^*}^2 +  \norm{x_{t} - x^*}^2  \right) .
\end{align*}
Similarly,
\begin{align*}
| \langle u^* - u_t, v^* - v_t \rangle | &\leq \mu \norm{u^* - u_t} \norm{v^* - v_t} \\
&\leq \frac{\mu}{2} \left( \norm{u^* - u_t}^2 + \norm{v^* - v_t} \right) \\
&= \frac{\mu}{2} \left( \norm{u^* + v^* - u_t - v_t}^2 \right) \\
&~~~~+ \mu | \langle u^* - u_t, v^* - v_t \rangle | ,
\end{align*}
which gives:
\[
| \langle u^* - u_t, v^* - v_t \rangle | \leq \frac{\mu}{2(1- \mu)} \norm{x^* - x_t}^2 .
\]
Combining, we get:
\begin{align*}
\mathbb{T}_2 &\leq \frac{\beta \mu}{2(1- \mu)} \left( 3\norm{x^* - x_t}^2 + \norm{x^* - x_{t+1}}^2 \right) .
\end{align*}
Moreover, by invoking RSC and Cauchy-Schwartz (similar to the proof of Theorem~\ref{thm:pgd}), we have:
\begin{align*}
\norm{x^* - x_t}^2 &\leq \frac{1}{\alpha} \left( F(x_t) - F(x^*) \right) + O(\varepsilon) ,\\
\norm{x^* - x_{t+1}}^2 &\leq \frac{1}{\alpha} \left( F(x_{t+1}) - F(x^*) \right) + O(\varepsilon) .
\end{align*}
Therefore we obtain the upper bound on $\mathbb{T}_2$:
\begin{align*}
\mathbb{T}_2 &\leq \frac{3 \beta \mu}{2 \alpha (1 - \mu)} \left( F(x_t) - F(x^*) \right) \\
&~~~ + \frac{\beta \mu}{2 \alpha (1 - \mu)} \left( F(x_{t+1}) - F(x^*) \right) + C' \varepsilon .
\end{align*}
Plugging in the upper bounds on $\mathbb{T}_1$ and $\mathbb{T}_2$ and re-arranging terms, we get:
\begin{align*}
&\left(1 - \frac{\beta \mu}{2\alpha (1 - \mu)} \right) (F(x_{t+1}) - F(x^*)) \\
&\leq \left(2 - \frac{\beta}{\alpha} + \frac{3 \beta \mu}{2 \alpha (1 - \mu)}  \right) (F(x_{t}) - F(x^*)) + C' \varepsilon,
\end{align*}
which leads to the desired result.
\endproof

\section{DISCUSSION}\label{sec:conc}

We provide some concluding remarks and potential directions for future work.

While our contributions in this paper are primarily theoretical, in recently published work~\cite{ganICASSP} we have explored the practical benefits of our approach in the context of linear inverse problems such as compressive sensing. However, our algorithms proposed in this paper are generic, and can be used to solve a variety of \emph{nonlinear} inverse problems. In future work, we will explore the empirical benefits for nonlinear settings, and also test the efficacy of our myopic PGD algorithm for handling model mismatch.

The main algorithmic message of this paper is to show that solving a variety of nonlinear inverse problems using a generative network model can be reduced to performing a set of $\varepsilon$-projections onto the range of the network model. This can be challenging in general; for most interesting generative networks, this itself is a nonconvex problem, and potentially hard. However, recent work by~\cite{hand-voroninski,heckel-hand} have studied special cases where this type of projection can be tractable; in particular, for certain neural networks satisfying certain randomness conditions, one can solve the projection problem using a variation of gradient descent (which is more or less what all approaches employ in practice). Studying the landscape of such projection problems is an interesting direction of future research.

We make several assumptions to enable our analysis. Some of them (for example, restricted strong convexity/smoothness; incoherence) are standard analysis tools and are common in the high-dimensional statistics and compressive sensing literature. However, in order to be applicable, they need to be verified for specific problems. A broader characterization of problems that \emph{do} satisfy these assumptions will be of great interest. 

\addtolength{\textheight}{-10cm}   



%

\section*{ACKNOWLEDGMENTS}

This project was supported in part by grants CAREER CCF-1750920 and CCF-1815101, a faculty fellowship from the Black and Veatch Foundation, and an equipment donation from the NVIDIA Corporation. The author would like to thank Ludwig Schmidt and Viraj Shah for helpful discussions.


\bibliographystyle{unsrt}
\bibliography{chinbiblio,vsbib,mrsbiblio}

\end{document}